\documentclass[a4paper,twoside]{article}

\usepackage{epsfig}
\usepackage{subcaption}
\usepackage{calc}
\usepackage{amssymb}
\usepackage{amstext}
\usepackage{amsmath}
\usepackage{amsthm}
\usepackage{multicol}
\usepackage{pslatex}
\usepackage{apalike}
\usepackage{algorithm2e}
\usepackage[bottom]{footmisc}

\usepackage{comment}  
\usepackage{bm}
\usepackage{multirow}
\usepackage{mathtools}  
\usepackage[per-mode=symbol]{siunitx}  
\usepackage[inline]{enumitem}  
\usepackage[hidelinks]{hyperref}

\usepackage{tikz}
\usepackage{varwidth}
\usetikzlibrary{arrows.meta, fit, positioning, shapes.geometric, calc}
\tikzset{
    block/.style={
        draw,
        fill=blue!20,
        rectangle,
        rounded corners,
        minimum height=2em,
        minimum width=5em,
        align=center,
        execute at begin node={\begin{varwidth}{7em}},
        execute at end node={\end{varwidth}}},
    sum/.style={
        draw,
        fill=blue!20,
        circle},
    square/.style={
        draw,
        regular polygon,
        regular polygon sides=4,
        fill=white,
    },
    input/.style={coordinate},
    output/.style={coordinate},
    pinstyle/.style={
        pin edge={
            Triangle[],
            thick,
            black}},
    arrow/.style={
        draw,
        thick,
        -{Triangle[]}},
    line/.style={
        draw,
        thick,
        -},
    triangle/.style={
        draw,
        fill=red!20,
        regular polygon,
        regular polygon sides=3}
}

\usepackage{listofitems} 

\DeclareMathAlphabet\mathbfcal{OMS}{cmsy}{b}{n}
\DeclareMathOperator{\diag}{diag}
\DeclareMathOperator*{\argmin}{arg\,min}
\newcommand{\expnumber}[2]{{#1} \cdot 10^{#2}}

\usepackage{xspace}
\def\ie{i.e.\@\xspace}
\def\wrt{w.r.t.\@\xspace}
\def\eg{e.g.\@\xspace}

\usepackage{SCITEPRESS}     

\begin{document}

\title{
Augmenting Neural Networks-based Model Approximators in Robotic Force-tracking Tasks
}

\author{
\authorname{
Kevin Saad\sup{1}\orcidAuthor{0009-0001-2295-0723}, 
Vincenzo Petrone\sup{2}\orcidAuthor{0000-0003-4777-1761},
Enrico Ferrentino\sup{2}\orcidAuthor{0000-0003-0768-8541},
Pasquale Chiacchio\sup{2}\orcidAuthor{0000-0003-3385-8866},\\
Francesco Braghin\sup{1}\orcidAuthor{0000-0002-0476-4118} and
Loris Roveda\sup{1,3}\orcidAuthor{0000-0002-4427-536X}
}
\affiliation{\sup{1}Department of Mechanical Engineering, Politecnico di Milano, 20133 Milano, Italy}
\affiliation{\sup{2}Department of Information Engineering, Electrical Engineering and Applied Mathematics (DIEM), University of Salerno, 84084 Fisciano, Italy}
\affiliation{\sup{3}Istituto Dalle Molle di Studi sull’Intelligenza Artificiale (IDSIA), Scuola Universitaria Professionale della Svizzera Italiana (SUPSI), Università della Svizzera Italiana (USI), 6962 Lugano, Switzerland}
\email{
kevin.saad@mail.polimi.it,\{vipetrone, eferrentino, pchiacchio\}@unisa.it,\\
francesco.braghin@polimi.it, loris.roveda@idsia.ch}
This work has been accepted for publication at ICINCO 2025 - 22nd International Conference on Informatics in Control,
Automation and Robotics on Thursday 3rd September, 2025.\\
Science and Technology Publications, Lda holds the copyright on the published version of this article.\\
Please refer to the published version in the conference proceedings at \url{https://www.doi.org/10.5220/0013830700003982   }
}


\abstract{
As robotics gains popularity, interaction control becomes crucial for ensuring force tracking in manipulator-based tasks.
Typically, traditional interaction controllers either require extensive tuning, or demand expert knowledge of the environment, which is often impractical in real-world applications. 
This work proposes a novel control strategy leveraging Neural Networks (NNs) to enhance the force-tracking behavior of a Direct Force Controller (DFC).
Unlike similar previous approaches, it accounts for the manipulator's tangential velocity, a critical factor in force exertion, especially during fast motions.
The method employs an ensemble of feedforward NNs to predict contact forces, then exploits the prediction to solve an optimization problem and generate an optimal residual action, which is added to the DFC output and applied to an impedance controller.
The proposed Velocity-augmented Artificial intelligence Interaction Controller for Ambiguous Models (VAICAM) is validated in the Gazebo simulator on a Franka Emika Panda robot.
Against a vast set of trajectories, VAICAM achieves superior performance compared to two baseline controllers.
}

\onecolumn \maketitle \normalsize \setcounter{footnote}{0} \vfill

\section{\uppercase{Introduction}}
\label{sec:introduction}

\begin{figure}
\centering
\begin{tikzpicture}

  \node[inner sep=0pt] (main) at (0,0) {
    \includegraphics[width=\columnwidth]{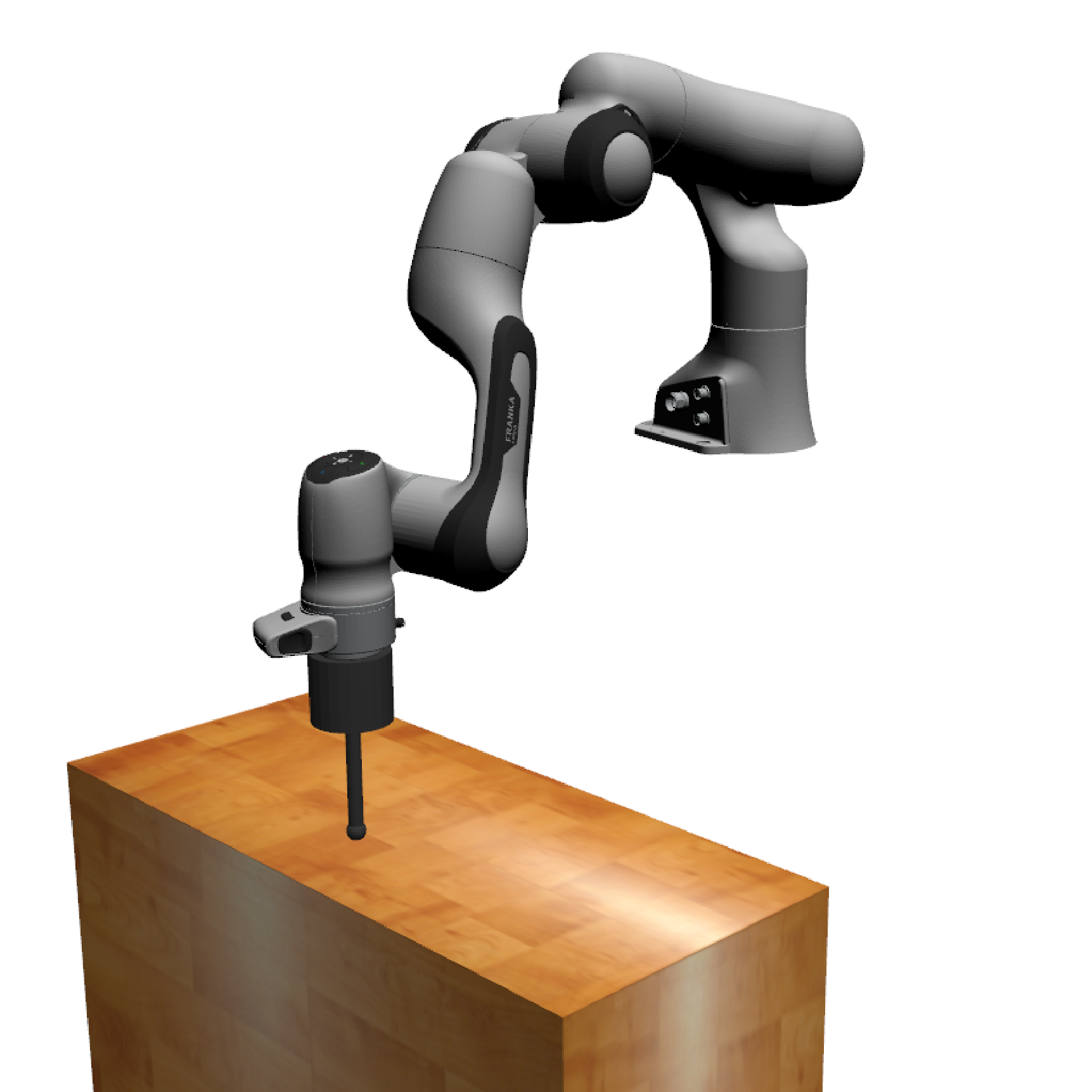}
  };

  \coordinate (startpoint) at ([xshift=0.325*\columnwidth,yshift=-0.76*\columnwidth]main.north west);

  \node[fill=white, inner sep=1pt, anchor=south west, draw=black, thick] (zoom) 
    at ([xshift=0.45\columnwidth, yshift=0.05\columnwidth]main.south west) {
    \includegraphics[width=0.5\columnwidth]{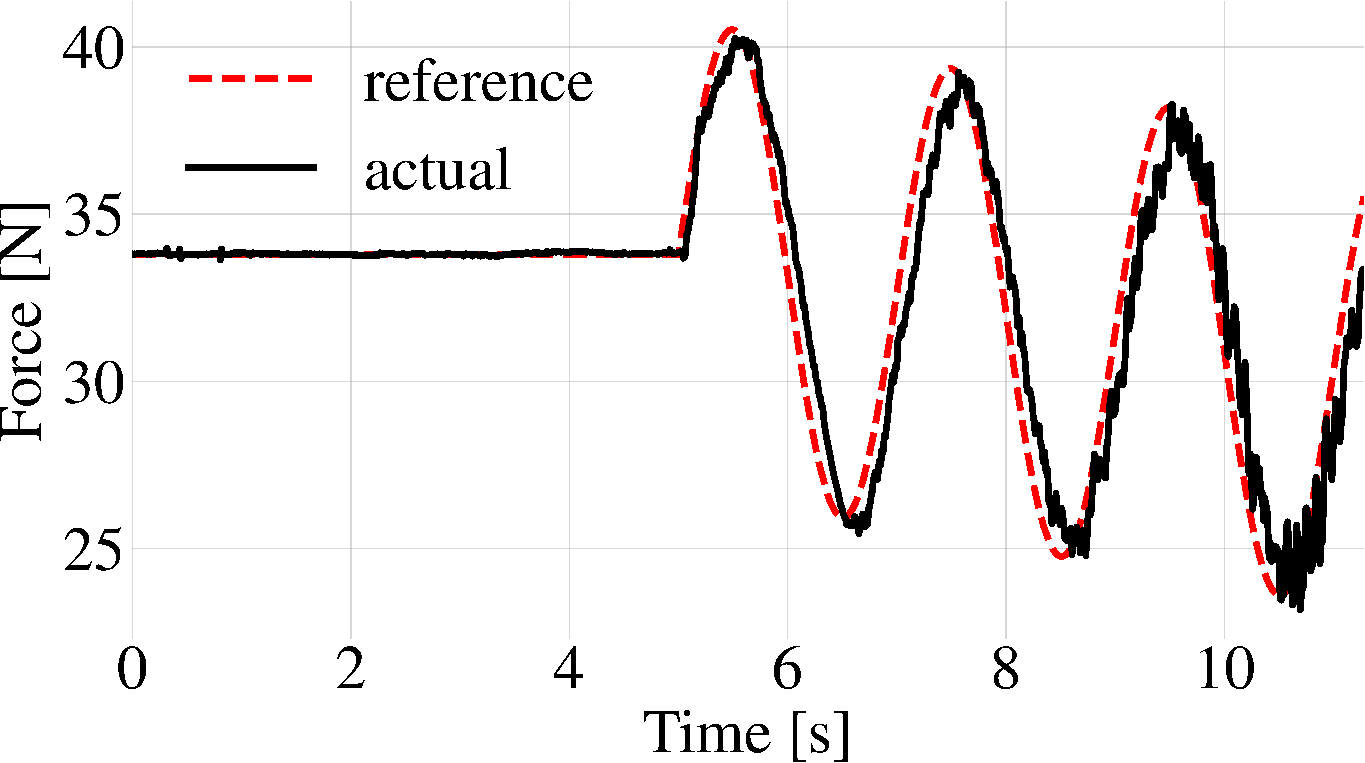}
  };

  \draw[dashed, thick] (startpoint) -- (zoom.north west);
  \draw[dashed, thick] (startpoint) -- (zoom.south west);

\end{tikzpicture}
\caption{Simulation setup --- the Panda robot performs a force-tracking task sliding on a wooden table with a spherical-tip end-effector.}
\label{fig:setup}
\end{figure}
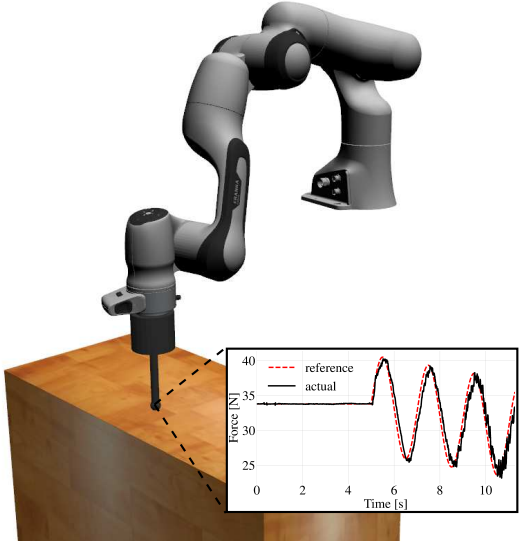

Modeling and controlling accurate force tracking in robotic manipulators remains a fundamental challenge for reliable robot-environment interaction. 
Achieving high force-tracking performance is critical in a broad spectrum of tasks, including contact-rich manipulation, precision assembly, and surface interaction (see Fig.~\ref{fig:setup}).

To address this challenge, impedance controllers achieve force-tracking accuracy through strategies such as reference generation \cite{roveda_sensorless_2021,huang_robust_2022,yu_adaptive_2024}, variable stiffness \cite{shen_fuzzy-based_2022,li_adaptive_2023}, and variable damping \cite{jung_force_2004,shu_adaptive_2021,duan_adaptive_2018,roveda_control_2020}.
Reference generation methods implement an explicit direct force control (DFC) loop to follow a desired force trajectory, while variable impedance approaches often rely on simplified linear-spring environment models \cite{jung_force_2004,shen_fuzzy-based_2022,li_adaptive_2023,shu_adaptive_2021,duan_adaptive_2018,yu_adaptive_2024}.
However, these simplified models typically represent only an approximation of the real environment dynamics \cite{roveda_sensorless_2021,jung_force_2004,matschek_safe_2023}.

To overcome modeling inaccuracies, recent literature has explored Artificial Intelligence (AI) techniques \cite{matschek_safe_2023}, particularly Neural Networks (NNs), to learn a model-less, data-driven mapping between the manipulator’s end-effector state and the exerted contact force directly at the control level.
In this context, ORACLE \cite{petrone_optimized_2025} was proposed as a controller that leverages NN-based models to optimize force tracking.
However, ORACLE’s original formulation neglects the effects of end-effector velocities tangential to the contact plane, potentially limiting its prediction accuracy and tracking performance, especially at high velocities.

This paper presents an extension of the ORACLE strategy by augmenting its model approximator's state with tangential velocity components, resulting in a refined controller named Velocity-Augmented Artificial Intelligence interaction Controller for Ambiguous Models (VAICAM).
The proposed approach constructs an accurate model of the environment that relates the end-effector pose, penetration velocity, and tangential velocity to the resulting contact forces using feed-forward neural networks (FFNNs) \cite{nagabandi_neural_2018}.

To this aim, a dedicated dataset containing dynamic trajectories in both force and position space is generated to train this model effectively.
An improved controller is then designed based on this augmented model, enabling optimal selection of control actions to minimize force-tracking errors.
Extensive validation is carried out in simulation, using Gazebo \cite{koenig_design_2004}, on a Franka Emika Panda manipulator \cite{haddadin_franka_2022}.
Furthermore, the paper conducts a comparative analysis between a standard direct force controller \cite{roveda_sensorless_2021} and ORACLE \cite{petrone_optimized_2025} across varying velocity conditions, concretely attesting ORACLE's performance degradation as the end-effector velocity increases.

By integrating tangential velocity information into the ORACLE framework, VAICAM demonstrates improved force prediction and enhanced tracking capabilities, extending the applicability of neural-network-based force controllers to a wider range of challenging interaction tasks.

\section{\uppercase{Methodology}}

This paper introduces VAICAM, an AI-driven tool to enhance the force tracking capabilities of an impedance controller used in unknown environments.
It augments the ensemble of FFNNs originally proposed in ORACLE \cite{petrone_optimized_2025} by adding to its input the tangential velocity $v$ of the end-effector (EE), resulting in a mapping from the robot state and the DFC control action ${\bm x}_f$ into its next state.
This design choice yields a more accurate prediction of the next wrench, that will later be used to compute the optimal residual action added to the low-level control action of the DFC.
VAICAM's overall control scheme is summarized in Fig.~\ref{fig:vaicam-architecture}, whose building blocks will be detailed in the next sections.

\definecolor{bleu}{rgb}{0.19,0.55,0.91}
\definecolor{alizarin}{rgb}{0.82,0.1,0.26}
\definecolor{amber}{rgb}{1.0,0.49,0.0}
\definecolor{ao}{rgb}{0.0,0.5,0.0}

\newlength{\hnodeDist}
\setlength{\hnodeDist}{1.5cm}
\newlength{\vnodeDist}
\setlength{\vnodeDist}{1.0cm} 

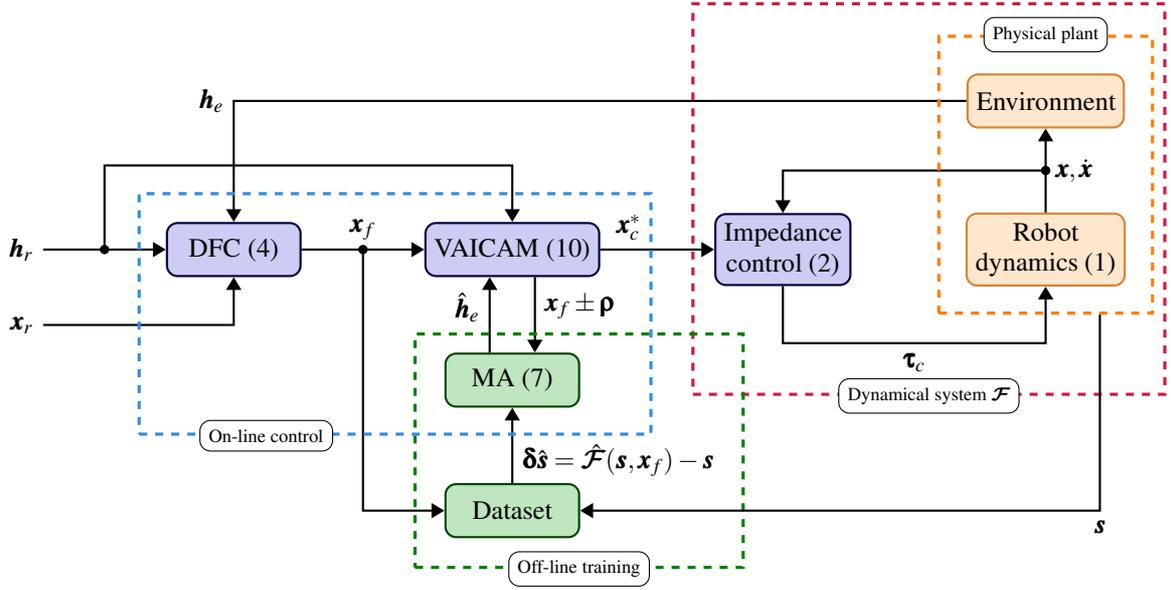
\begin{figure*}
    \centering
    \begin{tikzpicture}[auto,node distance=\vnodeDist and \hnodeDist,scale=1.0]
        \node [input](hr){};
        \node [sum, fill=black, scale=0.33, right=0.5\hnodeDist of hr](reference_wrench){};
        \node [input, below=of hr](xr){};
        \node [block, thick, draw=blue!80!black!30!black, fill=blue!80!black!20, right=0.5\hnodeDist of reference_wrench](dfc){DFC~\eqref{eq:methodology_DFC}};
        \node [sum, fill=black, scale=0.33, right=0.5\hnodeDist of dfc](action){};
        \node [block, thick, draw=blue!80!black!30!black, fill=blue!80!black!20, right=0.5\hnodeDist of action](vaicam){VAICAM~\eqref{eq:methodology_optimization_problem}};
        \node [block, thick, draw=blue!80!black!30!black, fill=blue!80!black!20, right=of vaicam](impedance_control){Impedance\\control~\eqref{eq:methodology_tau_ctrl}};
        \node [block, thick, draw=orange!80!black!80, fill=orange!20, right=of impedance_control, font=](robot){Robot\\dynamics~\eqref{eq:methodology_manipulator_model}};
        \node [sum, fill=black, scale=0.33, above=0.5\vnodeDist of robot](interim){};        
        \node [block, thick, draw=orange!80!black!80, fill=orange!20, above=0.5\vnodeDist of interim, font=](environment){Environment};
        \node [block, thick, draw=green!90!black!30!black,fill=green!60!black!25, below=of vaicam, font=](nn){MA~\eqref{eq:methodology_MA_transition_function}};
        \node [block, thick, draw=green!90!black!30!black,fill=green!60!black!25, below=of nn, font=](dataset){Dataset};

        \draw [arrow] (hr)--node[left,pos=0]{$\bm h_r$}(dfc);
        \draw [arrow] (dfc)--node{$\bm x_f$}(vaicam);
        \draw [arrow] (vaicam)--node[pos=0.25](mid_vaicam_impedance){$\bm x_c^*$}(impedance_control);
        \draw [arrow] (impedance_control.south)|-node[below,near end](control_torque){$\bm \tau_c$}([yshift=-0.75\vnodeDist]robot.south) -- (robot.south);
      
        \draw [arrow] (robot.north)--node[right]{$\bm x, \dot{\bm x}$}(environment.south);
        \draw [arrow] (environment.west)-|node[left]{$\bm h_e$}(dfc.north);
        \draw [arrow] (interim.west)-|(impedance_control.north);
        \draw [arrow] (xr)-|node[left,pos=0]{$\bm x_r$}(dfc);
        \draw [arrow] 
          let 
            \p1 = (vaicam.south),
            \p2 = ($(nn.north west)!2/3!(nn.north east)$)
          in
            (\x2,\y1) -- node[pos=0.4,right]{$\bm x_f \pm \bm\rho$} (\x2,\y2);
        \draw [arrow] 
          let 
            \p1 = ($(nn.north west)!1/3!(nn.north east)$),
            \p2 = (vaicam.south)
          in
            (\x1,\y1) -- node[pos=0.6,left]{$\hat{\bm h}_e$} (\x1,\y2);
        \draw [arrow] (action)|-(dataset.west);
        \draw [arrow] (dataset)--node[right,pos=0.3]{$\bm\delta \hat{\bm s}=\hat{\mathbfcal F}(\bm s, \bm x_f) - \bm s$}(nn);
        \draw [arrow] (reference_wrench)|-([yshift=0.75\vnodeDist]vaicam.north) -- (vaicam.north);

        \node[
            fit=(dfc)(vaicam)(nn)(mid_vaicam_impedance),
            draw,
            bleu,
            dashed,
            very thick,
            inner xsep=+5pt,
            xshift=-5pt,
            inner ysep=+7pt,
            yshift=-3pt,
        ](controlblock){};
        \path (controlblock.south west) -- (controlblock.south east) coordinate[pos=0.25] (controlblock_label_pos);
        \node[
            draw,
            rounded corners,
            fill=white,
        ] at (controlblock_label_pos) {\scriptsize On-line control};

        \node[
            fit=(robot)(environment),
            draw,
            amber,
            dashed,
            very thick,
            inner xsep=+10pt,
            inner ysep=12pt,
            yshift=2pt,
            label={[draw,rounded corners,fill=white,anchor=center,name=l]above: \scriptsize Physical plant}
        ](plant){};

        \node[
            fit=(impedance_control)(plant)(l)(control_torque),
            draw,
            alizarin,
            dashed,
            very thick,
            inner xsep=6pt,
            xshift=-2pt,
            inner ysep=+5pt,
            label={[draw, rounded corners, fill=white, anchor=center]below:\scriptsize Dynamical system $\bm{\mathbfcal F}$},
        ]{};

        \node[
            fit=(nn)(dataset),
            draw,
            ao,
            dashed,
            very thick,
            inner xsep=+36pt,
            xshift=25pt,
            inner ysep=9pt,
            yshift=-2pt,
            label={[draw, rounded corners, fill=white, anchor=center]below:\scriptsize Off-line training},
        ]{};

        \draw [arrow] ($(plant.south west)!0.75!(plant.south east)$) |- node{$\bm s$} (dataset.east);
               
    \end{tikzpicture}
    \caption{Control architecture. The MA learns the transition function $\bm{\hat{\mathbfcal F}}$ of the dynamical system represented by the impedance-controlled robot interacting with an unknown environment, given data composed of the system states $\bm s$ and control inputs $\bm x_f$, i.e. the DFC action. After training, VAICAM computes the optimal residual action $\bm x_c^*$, aiming at minimizing the force tracking error between $\bm h_r$ and the predicted wrench $\hat{\bm h}_e$.}
    \label{fig:vaicam-architecture}
\end{figure*}

\subsection{Base controller} \label{sec:methodology_base_controller}

The low-level impedance controller enforces the desired interaction dynamics on the robot, specifically, Cartesian space mass-spring-damper dynamics.
Consider the equations of motion for a manipulator with $n$ Degrees of Freedom (DOFs) performing an $m$-dimensional task, with $m \le 6 \le n$ \cite{featherstone_dynamics_2016}:
\begin{equation}\label{eq:methodology_manipulator_model}
     \bm B(\bm q)\ddot{\bm q} + \bm C(\bm q,\dot{\bm q})\dot{\bm q} + \bm\tau_{f}(\dot{\bm q}) + \bm g(\bm q) = \bm\tau_c - \bm J^\top(\bm q) \bm h_e,
\end{equation}
where $\bm B(\bm q) \in \mathbb R^{n \times n}$ is the inertia matrix, $\bm C(\bm q,\dot{\bm q}) \in \mathbb R^{n \times n}$ is the matrix accounting for the centrifugal and Coriolis effects, $\bm \tau_f(\dot{\bm q}) \in \mathbb R^n$ accounts for viscous and static friction, $\bm g(\bm q) \in \mathbb R^n$ represents the torque exerted on the links by gravity, $\bm \tau_c \in \mathbb R^n$ indicates the torque control action, $\bm J(\bm q) \in \mathbb R^{m \times n}$ is the geometric Jacobian, and $\bm h_e \in \mathbb R^m$ is the vector of wrenches exerted on the environment measured by means of a force/torque sensor mounted on the manipulator's flange.
The vectors $\bm q, \dot{\bm q}, \ddot{\bm q} \in \mathbb R^n$ represent joint positions, velocities, and accelerations, respectively.

The expression of the Cartesian impedance control law with robot dynamics compensation is \cite{siciliano_indirect_1999,formenti_improved_2022}
\begin{equation}\label{eq:methodology_tau_ctrl}
    \bm\tau_c = \bm J^\top(\bm q) \bm h_c + \bm C(\bm q,\dot{\bm q})\dot{\bm q} + \bm\tau_f(\dot{\bm q}) + \bm g(\bm q),
\end{equation}
where the task space wrench $\bm h_c$ realizing the compliant behavior can be chosen as \cite{caccavale_six-dof_1999,iskandar_hybrid_2023}
\begin{equation}\label{eq:methodology_h_c_simplified}
    \bm h_c = \bm K_d \bm\Delta \bm x + \bm D_d \dot{\bm x},
\end{equation}
where $\bm K_d, \bm D_d \in \mathbb R^{m \times m}$ are diagonal matrices of control parameters, namely stiffness and damping, respectively, and $\bm\Delta \bm x \triangleq \bm x_d - \bm x \in \mathbb R^m$ is the Cartesian pose error between the setpoint $\bm x_d \in \mathbb R^m$ and the actual robot pose $\bm x \in \mathbb R^m$.
Assuming $m = 6$, $\bm x$ is defined as $ \bm x \triangleq (x, y, z, \phi, \theta, \psi)^\top$, where $(x,y,z)^\top$ and $(\phi,\theta,\psi)^\top$ are translational and rotational components, respectively.

\subsection{Direct Force Controller}

Given that the impedance controller solely manages interaction forces passively, lacking the capability to track a force reference, a DFC loop can be closed specifically along the directions in which force tracking is necessary \cite{roveda_sensorless_2021}.
The adopted control law is a simple PI controller having the following model:
\begin{equation}\label{eq:methodology_DFC}
    \bm x_f =\bm x_r + \bm\Gamma \left( \bm K_P \bm\Delta \bm h + \bm K_I \int_t{\bm\Delta \bm h dt} \right), 
\end{equation}
where, if $m=6$, $\bm\Gamma = \diag (\gamma_x, \gamma_y, \gamma_z, \gamma_\phi, \gamma_\theta, \gamma_\psi)$ is the task specification matrix \cite{khatib_unified_1987}, with $\gamma_i = 1$ if the $i$-th direction is subject to force control, $0$ otherwise.
$\bm x_f \in \mathbb R^m$ is the force controller output, while $\bm x_r \in \mathbb R^m$ is the reference pose, whose $i$-th component is tracked when $\gamma_i = 0$.
$\bm K_P, \bm K_I \in \mathbb R^{m \times m}$ are the proportional and integral gains of the controller, and $\bm\Delta \bm h = \bm h_r - \bm h_e \in \mathbb R^m$ is the error between the reference wrench to be exerted $\bm h_r \in \mathbb R^m$ and the actual exerted wrench $\bm h_e \in \mathbb R^m$.

\subsection{Model Approximator}\label{sec:methodology_ma}

The Model Approximator (MA) addresses the inherent complications in accurately modeling the robot-environment interaction with a rather straightforward method that only requires the user to set up a handful of experiments that autonomously train the NN-based model.
The MA deals with the current system state $\bm s_k$, aiming to predict the state at the next time step $k+1$ following the equation
\begin{equation}\label{eq:methodology_dynamics_transtion_function_approximate}
    \hat{\bm s}_{k+1} = \hat{\mathbfcal F}(\bm s_k, \bm x_f),
\end{equation}
where $\hat{\bm s}_{k+1}$ is the predicted next state.
$\hat{\mathbfcal F}$ represents the transition dynamics approximation which outputs the next predicted state, where $\mathbfcal F$ indicates the actual dynamical system, \ie the impedance-controlled robot (see Fig.~\ref{fig:vaicam-architecture}).
Specifically, instead of predicting $\bm s_{k+1}$ explicitly, the actual FFNN's output $\bm\delta \hat{\bm s}$ is chosen to be the approximate difference between two subsequent states, similarly to \cite{nagabandi_neural_2018}, \ie:
\begin{equation}\label{eq:methodology_delta_s}
    \bm\delta \bm s = \bm s_{k+1} - \bm s_k.
\end{equation}
This allows \eqref{eq:methodology_dynamics_transtion_function_approximate} to be rewritten as
\begin{equation}\label{eq:methodology_MA_transition_function}
    \hat{\bm s}_{k+1} = \bm s_k + \bm\delta \hat{\bm s}(\bm s_k, \bm x_f),
\end{equation}
with $\bm\delta \hat{\bm s}(\bm s_k, \bm x_f)$ being the actual NN output, as in Fig.~\ref{fig:ffnn}.


\tikzset{>=latex} 

\colorlet{myred}{red!80!black}
\colorlet{myblue}{blue!80!black}
\colorlet{mygreen}{green!60!black}
\colorlet{myorange}{orange!70!red!60!black}
\colorlet{mydarkred}{red!30!black}
\colorlet{mydarkblue}{blue!40!black}
\colorlet{mydarkgreen}{green!30!black}

\tikzset{
  >=latex, 
  node/.style={thick,circle,draw=myblue,minimum size=17,inner sep=0.5,outer sep=0.6},
  node in/.style={node,rectangle,rounded corners=8,green!20!black,draw=mygreen!30!black,fill=mygreen!25,inner sep=2pt, outer sep=0pt},
  node hidden/.style={node,blue!20!black,draw=myblue!30!black,fill=myblue!20},
  node out/.style={node,red!20!black,draw=myred!30!black,fill=myred!20},
  connect/.style={thick,mydarkblue}, 
  connect arrow/.style={thick, -{Triangle[]}},
  node 1/.style={node in}, 
  node 2/.style={node hidden},
  node 3/.style={node out}
}
\def\nstyle{int(\lay<\Nnodlen?min(2,\lay):3)} 

\newcommand\setAngles[3]{
  \pgfmathanglebetweenpoints{\pgfpointanchor{#2}{center}}{\pgfpointanchor{#1}{center}}
  \pgfmathsetmacro\angmin{\pgfmathresult}
  \pgfmathanglebetweenpoints{\pgfpointanchor{#2}{center}}{\pgfpointanchor{#3}{center}}
  \pgfmathsetmacro\angmax{\pgfmathresult}
  \pgfmathsetmacro\dang{\angmax-\angmin}
  \pgfmathsetmacro\dang{\dang<0?\dang+360:\dang}
}

\begin{figure}
\small
\centering
\begin{tikzpicture}[x=1.3cm,y=0.8cm]
  \message{^^JNeural network, shifted}
  \readlist\Nnod{2,3,3,3,1} 
  \readlist\Nstr{n,m,m,m,k} 
  \readlist\Cstr{\strut x,a^{(\prev)},a^{(\prev)},a^{(\prev)},y} 
  \def\yshift{0.6} 
  
  \message{^^J  Layer}
  \foreachitem \N \in \Nnod{ 
    \def\lay{\Ncnt} 
    \pgfmathsetmacro\prev{int(\Ncnt-1)} 
    \message{\lay,}
    
    \foreach \i [evaluate={
                    \c=int(\i==\N || (\i==1 && \lay==1) ? 1 : 0);
                    \x=\lay;
                    \y=\N/2-\i-\c*\yshift;
                    \n=\nstyle;}] in {1,...,\N}{ 
      \ifnum\lay=1  
          \ifnum\i=1  
            \node[node \n] (N\lay-\i) at (\x,\y) {$\bm x_f$};
          \else  
            \node[node \n] (N\lay-\i) at (\x,\y) {$\bm{s}_k$ \eqref{eq:ma_state}};
          \fi
      \else
          \ifnum\lay=\Nnodlen  
            \node[node \n] (N\lay-\i) at (\x,\y) {$\bm \delta \hat{\bm s}$};
          \else  
            \node[node \n] (N\lay-\i) at (\x,\y) {};
          \fi
      \fi
    
      \ifnum\lay>1 
        \foreach \j in {1,...,\Nnod[\prev]}{ 
          \draw[connect,white,line width=1.2] (N\prev-\j) -- (N\lay-\i);
          \draw[connect] (N\prev-\j) -- (N\lay-\i);
        }
      \fi 
    }

    \ifnum\lay>1 \ifnum\N>1
      \path (N\lay-\N) --++ (0,1+\yshift) node[midway,scale=1.5] {$\vdots$};
    \fi\fi
  }
  
  \node[above of=5,node distance=0.5cm,align=center,mygreen!60!black] at (N1-1.90) {input\\[-0.2em]layer};
  \node[above of=2,node distance=0.25cm,align=center,myblue!60!black] at (N3-1.90) {hidden layers $\hat{\mathbfcal F}$};
  \node[above of=10,node distance=0.5cm,align=center,myred!60!black] at (N\Nnodlen-1.90) {output\\[-0.2em]layer};

  \node[node, align=center, black, right=0.75cm of N\Nnodlen-1] (sum) {$+$};
  
  \node[coordinate, below of=N3-3, node distance=0.5cm] (ghost) {};

  \draw[connect, black] (N1-2) |- (ghost);
  \draw[connect arrow, black] (ghost) -| (sum);

  \draw[connect arrow, black] (N\Nnodlen-1) -- (sum);

  \node[coordinate, right=0.75cm of sum] (output) {};
  \draw[connect arrow, black] (sum) -- node[below] {$\hat{\bm s}_{k+1}$} (output);
\end{tikzpicture}
\caption{MA architecture. The hidden layers approximate the dynamics transition function $\bm s_{k+1} = \mathbfcal F(\bm s_k, \bm x_f)$ by predicting the state variation $\bm\delta \hat{\bm s}$.}
\label{fig:ffnn}

\end{figure}
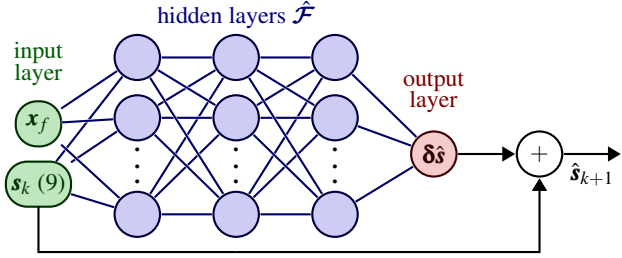

As regards the state definition, in general $\bm s$ takes the form
\begin{equation} \label{eq:ma_state_general}
\bm s \triangleq (\bm x, \dot{\bm x}, \bm h_e)^\top,
\end{equation}
but it might be specialized according to the task setup.
Indeed, assuming tracking forces only along the $z$ axis, ORACLE \cite{petrone_optimized_2025} only considers orthogonal components in $\bm s$, \ie EE penetration velocity and force along the same direction.
This work introduces a new feature in $\bm s$, namely the EE tangential velocity $v$, which influences the evolution of the exerted force, especially at high speeds \cite{iskandar_hybrid_2023}.
In summary, we propose the following augmentation in the MA state representation:
\begin{equation}\label{eq:ma_state}
    \bm s \triangleq \left( z, \dot z, f_z \right)^\top \rightarrow \bm s \triangleq \left( z, \dot z, v, f_z \right)^\top,
\end{equation}
where $v = \sqrt{\dot x^2 + \dot y^2}$ is the tangential velocity, with $\dot x, \dot y$ being the velocity components on the EE $xy$ plane, and $f_z$ is the normal force along $z$, \ie, the contact direction.

From an architectural standpoint, the FFNN is actually an \textit{ensemble} of FFNN as in \cite{chua_deep_2018}, since using an array of $N$ independently trained NNs minimizes the errors risen by epistemic and stochastic uncertainties.

\subsection{VAICAM Algorithm}\label{sec:method_VAICAM}

Acting as the glue that joins all items together, the VAICAM algorithm combines the DFC-enhanced impedance controller and the MA.
It utilizes the control action output by the DFC $\bm x_f$ along with the force prediction of the MA $\hat{\bm h}_e$, and then outputs the optimal residual action $\bm x_c^*$ that will then be added to the impedance controller action aiming at minimizing the force tracking error.
VAICAM will search for $\bm x_c^*$ in the neighborhood of $\bm x_f$ in an area centered around it with predefined radius $\bm\rho > \bm 0$.

Then, the control input to the impedance controller is updated by solving an optimization problem, run at each control step $k$:
\begin{equation}\label{eq:methodology_optimization_problem}
   \bm x_c^*(k) = \argmin_{ \bm x_c \in \bm x_f \pm \bm \rho} \mathcal L_k(\bm s, \bm x_c, \bm h_r), 
\end{equation}
where
\begin{equation}\label{eq:methodology_cost_function_optimizer}
    \mathcal L_k(\bm s, \bm x_c, \bm h_r) = \lvert \bm h_r(k) -  \hat{\bm h}_e(\bm s_k, \bm x_c) \rvert + \Omega_k(\bm x_c)
\end{equation}
is the cost function to minimize, whose first term $\lvert \bm h_r - \hat{\bm h}_e \rvert$ is the expected force tracking error, and
\begin{equation}\label{eq:methodology_omega_k}
    \Omega_k(\bm x_c) = \lVert \bm x_c \rVert_{\bm \alpha}^2 + \lvert \bm x_c - \bm x_c^*(k-1) \rvert_{\bm \beta}
\end{equation}
is a regularizer that contributes to smoothing-out large jumps of the control term $\bm x_c$.
The first term in \eqref{eq:methodology_omega_k} penalizes heavy actions so as to avoid deep penetrations of the EE into the environment, whereas the second prevents fast variations between subsequent actions. 
The two terms are defined as:
\begin{subequations}\label{eq:methodology_regularizer}
\begin{align}
    \lVert \bm x_c \rVert_{\bm \alpha}^2 & = \sum_i{\alpha_i x_{c,i}^2},\label{eq:methodology_regularizer_left} \\ 
    \lvert \bm x_c - \bm x_c^*(k-1) \rvert_{\bm \beta} & = \sum_i{\beta_i \lvert x_{c,i} - x^*_{c,i}(k-1) \rvert},
\end{align}
\end{subequations}
where both $\bm\alpha, \bm\beta \in \mathbb R^m$ are parameters to be chosen by the user.
This method, summarized in Algorithm~\ref{alg:methodology_voracle_algorithm}, along with the base force controller, is activated as soon as contact is established between the EE and the environment.

\begin{algorithm}[t]
\caption{VAICAM algorithm at every control step $k$}
\label{alg:methodology_voracle_algorithm}

\small

\KwData{$\bm x_f$ from \eqref{eq:methodology_DFC}, $\bm s = (\bm x, \dot{\bm x}, \bm h_e)^\top, \bm h_r$}
\KwResult{$\bm x^*_c$}
\If{$k = 1$}{
    Set $\bm\alpha$, $\bm\beta$ and $\bm\rho$\;
    Set weights of the pre-trained MA NN ensemble\;
}
Build the discretized neighborhood of candidate optimal actions $\bm B_{\bm\rho}(\bm x_f) = [ \bm x_f \pm \bm\rho ]$\;
\For{$\bm x_c \in \bm B_\rho(\bm x_f)$}{
Infer the predicted force from the model approximator with \eqref{eq:methodology_MA_transition_function}: $\hat{\bm h}_e (\bm s, \bm x_c) = \bm h_e + \bm\delta\hat{\bm h}_e(\bm s, \bm x_c)$\;
Compute the regularizer in \eqref{eq:methodology_omega_k} as $\Omega_k(\bm x_c) = \lVert \bm x_c \rVert_{\bm\alpha}^2 + \lvert \bm x_c - \bm x_c^*(k-1) \rvert_{\bm\beta}$\;
Compute the corresponding loss function in \eqref{eq:methodology_cost_function_optimizer} as $\mathcal L_k(\bm s, \bm x_c, \bm h_r) = \lvert \bm h_r - \hat{\bm h}_e \rvert + \Omega_k(\bm x_c)$\;
}
Call VAICAM by solving the optimization problem in \eqref{eq:methodology_optimization_problem} minimizing \eqref{eq:methodology_cost_function_optimizer}: $\bm x_c^*(k) = \argmin
_{\bm x_c \in \bm B_{\bm\rho}(\bm x_f)}
{\mathcal L_k
(\bm s, \bm x_c, \bm h_r)
}$\;

\end{algorithm}

\subsection{Training Procedure} \label{sec:methodology_training_procedures}

The training and validation of the ensemble of FFNN, which is done in a preliminary offline stage, uses collected data resulting from thorough exploration of the state space.
In order to collect the data, reference forces and positions are provided to the base force controller, and the output data, consisting of the actual robot states $\bm s$ and control actions $\bm x_f$, are recorded.
During the training stage of the FFNN, their weights are updated using the Stochastic Gradient Descent algorithm in order to minimize the Mean Squared Error (MSE) between the actual and estimated states.

In our training procedure, we recommend commanding sine waves as references in both force and position spaces, in order to have a thorough exploration and avoid data gaps in the state domain.
Commanding a sine wave reference for the position additionally entails that the EE velocity $v$ is sinusoidal, allowing for the complete exploration of the velocity space as well.
This is a crucial aspect because, compared to \cite{petrone_optimized_2025}, the inclusion of the new feature $v$ requires a dedicated training. 
Furthermore, we recommend exaggerating the amplitude of the sine wave force references: even though they cannot be perfectly tracked by the controller, this ensures that the force domain is sufficiently explored.

After collecting data, they are then processed following methods used in \cite{nagabandi_neural_2018,chua_deep_2018}, \ie they are normalized by subtracting the mean of each quantity and then dividing by its standard deviation.
A zero-mean Gaussian noise in the form $\mathcal N(\mu = 0, \sigma)$ is applied to the measured data $\bm h_e$ in order to enhance the robustness of the NN.

\section{\uppercase{Results}}

\subsection{Task and Materials}

The experimental validation is divided into two main phases, both conducted on the 7-DOF Franka Emika Panda robot:
\begin{itemize}
\item \textbf{Experiment I:} train, validate, and test the Static Model Approximator (SMA) used by ORACLE, which is the MA trained and validated using \emph{static} position references \cite{petrone_optimized_2025} \emph{without} tangential velocity, and the Dynamic Model Approximator (DMA) used by VAICAM, which is the MA trained and validated using \emph{dynamic} position references \emph{with} tangential velocities;
    \begin{itemize}
    \item the goal is to assess the performance of both MAs on dynamic position reference trajectories, and validate that the DMA yields higher accuracy as the tangential velocity increases;
    \end{itemize}
\item \textbf{Experiment II:} execute dynamic trajectories using the base controller \cite{roveda_sensorless_2021}, ORACLE \cite{petrone_optimized_2025} and VAICAM, and compare the force tracking results;
    \begin{itemize}
    \item in this case, the objective is to assess the performance of both control strategies, and validate that the controller that uses the DMA (VAICAM) performs better than both the base controller and the controller that uses the SMA (ORACLE).
    \end{itemize}
\end{itemize}

The NN's algorithm is coded in Python using PyTorch \cite{paszke_pytorch_2019}, interfacing with the other modules via ROS communication mechanisms \cite{quigley_ros_2009}.
The controllers implemented in ROS are coded in C++.

In order to accurately simulate the robot, Gazebo \cite{koenig_design_2004} is used as the simulation software.
A spherical tip EE is mounted at the flange of the robot (see Fig.~\ref{fig:setup}).
The interaction control parameters are chosen as $\bm K_d = \diag{(K_{d,t}, K_{d,t}, K_{d,t}, K_{r,t}, K_{r,t}, K_{r,t})}$ and $\bm D_d = \diag{\{\xi \sqrt{K_{d,i}}\}_{i=1}^6}$, where $K_{d,t}$ and $K_{d,r}$ are the translational and rotational stiffness gains, respectively, and $\xi$ is the damping ratio.
Table~\ref{table:simulated_environment_parameters} provides a summary of the parameters.
In our experiment, a linear force is exerted on the $z$ axis, thus $\bm\Gamma = \diag{(0,0,1,0,0,0)}$.

\begin{table}
\centering
\caption{Parameters used in the experiments.}
\label{table:simulated_environment_parameters}
\resizebox{1.0\columnwidth}{!}{
\begin{tabular}{|c|c|c|}
\hline
\textbf{Parameter} & \textbf{Value} \\ 
\hline
Coulomb friction coefficient $\mu$ & 0.2 \\
Time step solver & ODE \\
Impedance control translational stiffness $K_{d,t}$ & 1700 \\
Impedance control orientational stiffness $K_{d,r}$ & 300 \\
Damping ratio $\xi$ & 1 \\
DFC Proportional gain $K_P$ & $\expnumber{1}{-6}$ \\
DFC Integral gain $K_I$ & $\expnumber{2}{-3}$ \\
\hline
\end{tabular}
}
\end{table}

\subsection{Experiment I: Model Approximator Validation} \label{sec:results_sim_model_approximator}

\begin{table}
\begin{center}
\caption{Configurations of the FFNN ensemble.}
\label{table:FFNN_configuration}
\begin{tabular}{|c|c|c|} 
\hline
\textbf{Parameter} & \textbf{Value} \\ 
\hline
Number of estimators $N$ & 3\\
Hidden layers & 3\\
Neurons per layer & 200\\
Activation function & ReLU\\
Learning rate & $\expnumber{1}{-3}$\\
Ensemble type & Fusion\\
Training epochs & 50\\
Loss function & MSE\\
Weight optimizer & Adam\\
\hline
\end{tabular}
\end{center}
\end{table}

For a fair comparison with \cite{petrone_optimized_2025}, we base the FFNNs structure on the findings therein.
Specifically, all FFNNs used share the same configuration in terms of depth (number of hidden layers), width (number of neurons per layer) and learning algorithm parameters, as listed in Table~\ref{table:FFNN_configuration}.
Linear type layers are used, while the activation function is of the ReLU type \cite{agarap_deep_2018}.
The optimized configuration of the base estimator that ensures enhanced inference time without compromising prediction accuracy uses $N = 3$ FFNNs.
Adam \cite{kingma_adam_2015} refers to Adaptive Moment Estimation, an optimizer used for NN regression tasks, while ``Fusion'' indicates how a single output is retrieved from the ensemble, \ie the arithmetic average across the independent network estimations is computed.

\subsubsection{Static Model Approximator} \label{sec:results_sim_static_ma}

The dataset is composed of 10 trajectories, with 9 of them being used in the training set, and the remaining 1 in the validation set, used by Adam \cite{kingma_adam_2015} to adaptively optimize the learning rate.
The position reference is in the form of a static waypoint, as this MA, unlike the one proposed in this work, does not take into consideration $v$ \cite{petrone_optimized_2025}.
The force reference is in the form of a sinusoidal wave with randomized frequency, amplitude, and mean.
Raw data in both training and validation sets are pre-processed as indicated in Sect.~\ref{sec:methodology_training_procedures}. 

After training the MA, a test set is developed in order to assess its performance.
Since the aim of this work is to assess the MA generalization ability against dynamic position trajectories, this set is composed of horizontal line references that the EE tries to track with a constant velocity and a sinusoidal force reference.
The MA is tested against a total of 110 trajectories of about $\SI{1.2}{\meter}$ executed using the base controller, divided into 11 evenly-spaced velocities in the range $[0.01, 0.50]\,\si{\meter\per\second}$.
The performance assessment is based on the comparison of the force predicted by the MA with the actual force.

\subsubsection{Dynamic Model Approximator} \label{sec:sim_DMA}

Compared to the SMA discussed in Sect.~\ref{sec:results_sim_static_ma}, we adopt a different training approach, \ie both the force and position references in the training set take a sinusoidal form.
On the basis of the training procedure outlined in Sect.~\ref{sec:methodology_training_procedures}, sinusoidal position references allow covering the desired velocity span better than constant velocity profiles.
The validation set is composed of 11 trajectories, each randomly given a velocity from the 11 velocity samples in the range $[0.01, 0.50]\,\si{\meter\per\second}$.

After executing the trajectories, the data stored in the dataset are pre-processed according to the procedure reported in Sect.~\ref{sec:methodology_training_procedures}.
Using the DMA, the network now takes into consideration $v$ while predicting the next state, which ameliorates the performance of the MA at higher velocities.
In order to confirm this thesis, the DMA is evaluated on the same test set as the SMA's, \ie using the same 110 line trajectories.

\subsubsection{Model Approximators Comparison}

\begin{figure}
\centering
\includegraphics[width=\columnwidth]{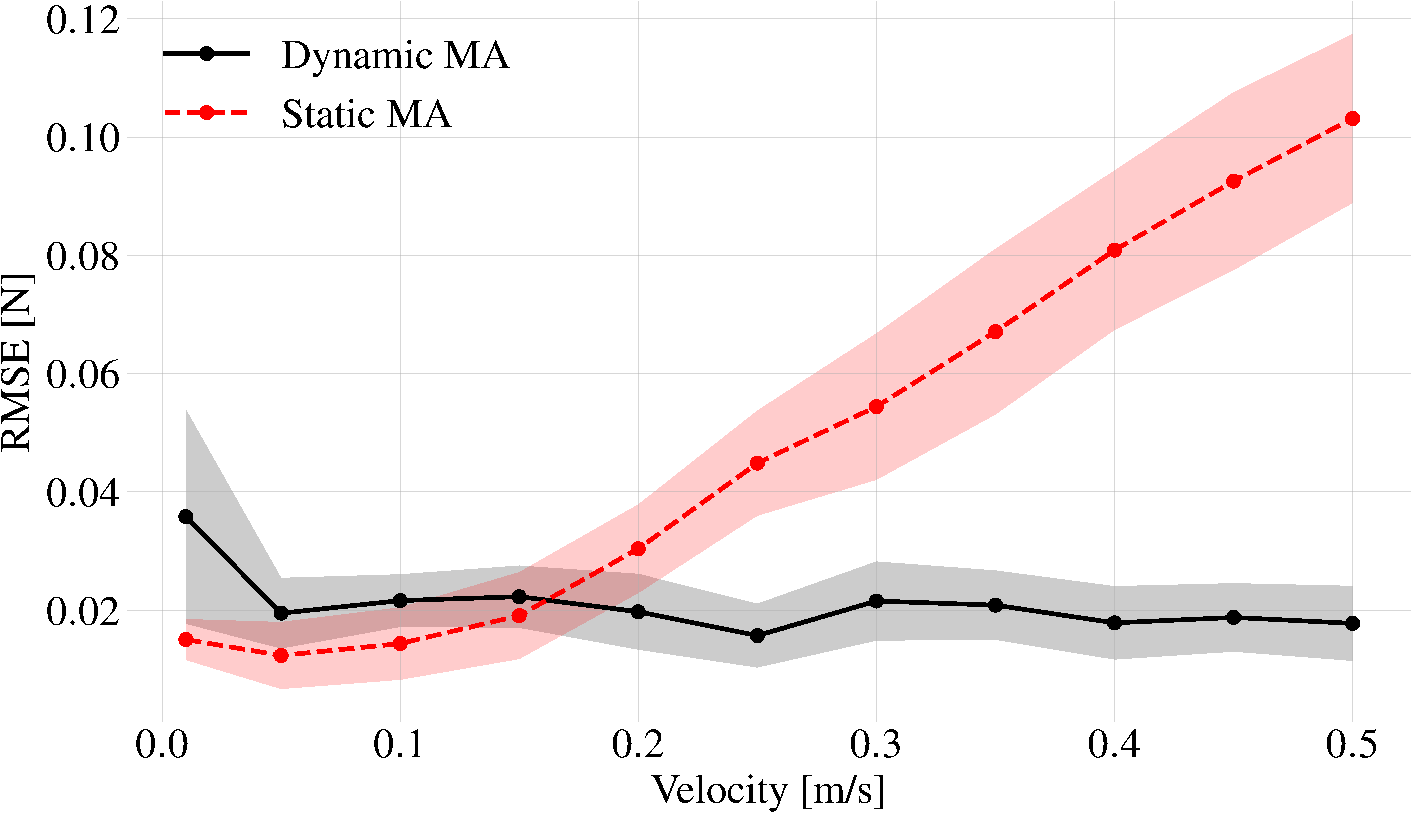}
\caption{Comparison between the static and the dynamic model approximators, in terms of RMSE, as the EE velocity increases.}
\label{fig:Average_MSE_vs_Velocity_of_Static_and_Dynamic_MA}
\end{figure}

Fig.~\ref{fig:Average_MSE_vs_Velocity_of_Static_and_Dynamic_MA} reports the Root Mean Square Error (RMSE) between the predicted and measured force, when either the SMA or DMA is employed, for increasing EE tangential velocity $v$.
As expected, the plot reveals a better generalization ability of the DMA over the SMA to higher EE velocities.
While the results are comparable at low speed, as $v$ increases the SMA showcases lower prediction accuracy, in terms of both mean and variance across trajectories.

This qualitative analysis is quantitatively confirmed by Table \ref{table:ma}, which also reports the numerical improvement factor $\eta$ of DMA over SMA.
As $v$ increases, the former outperforms the latter by up to 454\%, in terms of average RMSE recorded across trajectories at the same velocity.

\begin{table}
\centering
\caption{Average RMSE of static and dynamic model approximators across trajectories --- $\eta$ indicates the improvement factor of DMA over SMA.}
\label{table:ma}
\begin{tabular}{|c|c|c|c|} 
\hline
\textbf{Velocity} & \multicolumn{2}{c|}{\textbf{RMSE [\si{\newton}] }} & \multirow{2}{*}{$\bm\eta$} \\
\textbf{[\si{\meter\per\second}]} & \multicolumn{1}{c}{\textbf{SMA}} & \textbf{DMA} &  \\ 
\hline
0.01 & 0.0154 & 0.0397 & 0.3877 \\
0.05 & 0.0135 & 0.0203 & 0.663 \\
0.1 & 0.0155 & 0.022 & 0.7029 \\
0.15 & 0.0203 & 0.0228 & 0.8893 \\
0.2 & 0.0312 & 0.0207 & 1.511 \\
0.25 & 0.0456 & 0.0165 & 2.7628 \\
0.3 & 0.0557 & 0.0225 & 2.4777 \\
0.35 & 0.0684 & 0.0216 & 3.1683 \\
0.4 & 0.0819 & 0.0188 & 4.3528 \\
0.45 & 0.0936 & 0.0196 & 4.7841 \\
0.5 & 0.104 & 0.0188 & 5.5422 \\
\hline
\end{tabular}
\end{table}

\subsection{Experiment II: Control Algorithm Validation} \label{sec:results_sim_controlalg}

Once the SMA and the DMA are trained, the three control algorithms --- DFC \cite{roveda_sensorless_2021}, ORACLE \cite{petrone_optimized_2025}, and VAICAM (ours) --- can be tested against a test set aiming to prove that VAICAM outperforms its competitors.
The optimizer parameters are tuned as $\alpha = 25$, $\beta = 200$, and $\rho = \SI{0.003}{\meter}$ for both ORACLE and VAICAM.

\begin{figure}
\centering
\includegraphics[width=\columnwidth]{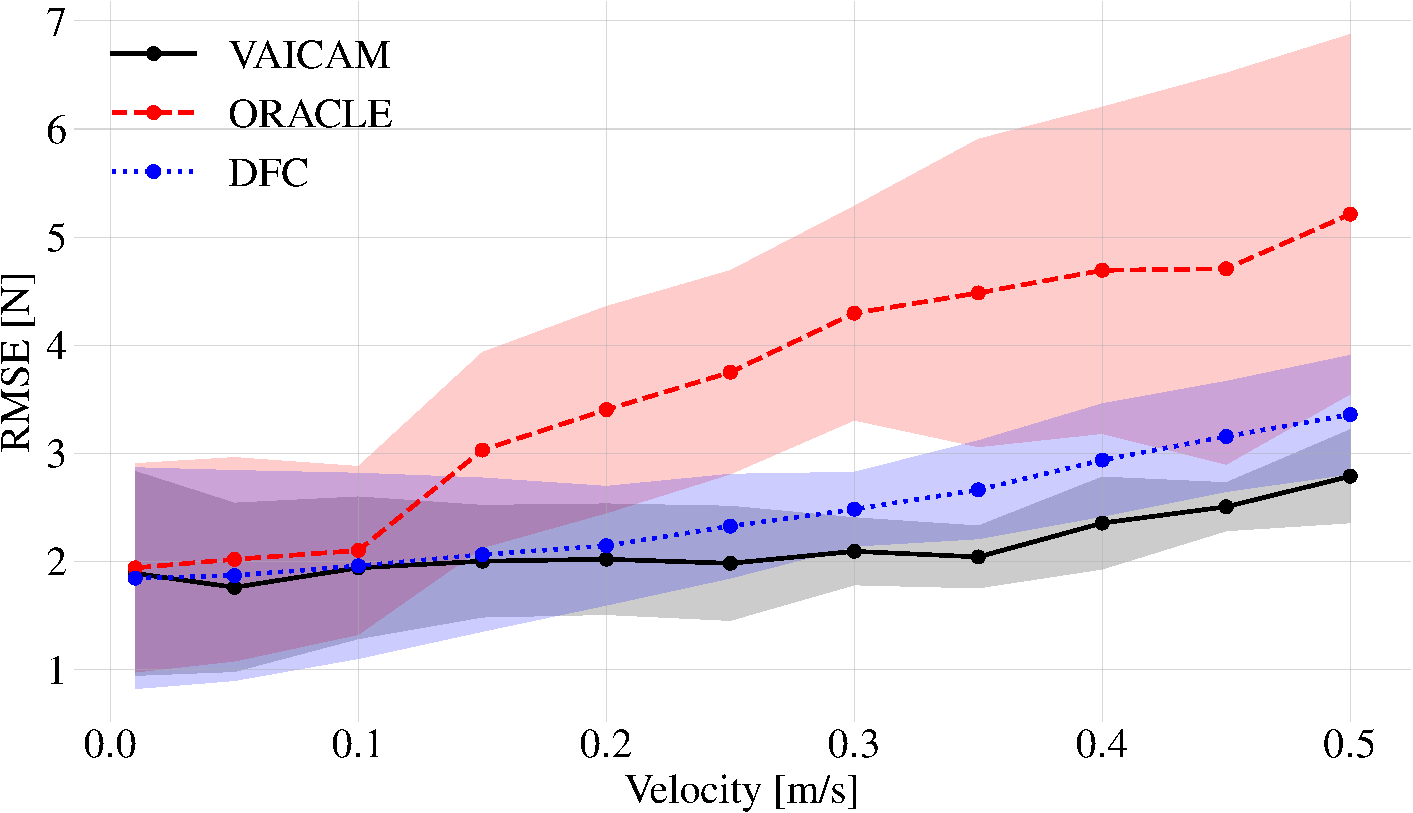}
\caption{Comparison between the baseline DFC \cite{roveda_sensorless_2021}, ORACLE \cite{petrone_optimized_2025}, and VAICAM (ours), in terms of RMSE, as the EE velocity increases.}
\label{fig:oracle_vaicam_dfc_mse_sim}
\end{figure}

\begin{figure}
\centering
\includegraphics[width=\columnwidth]{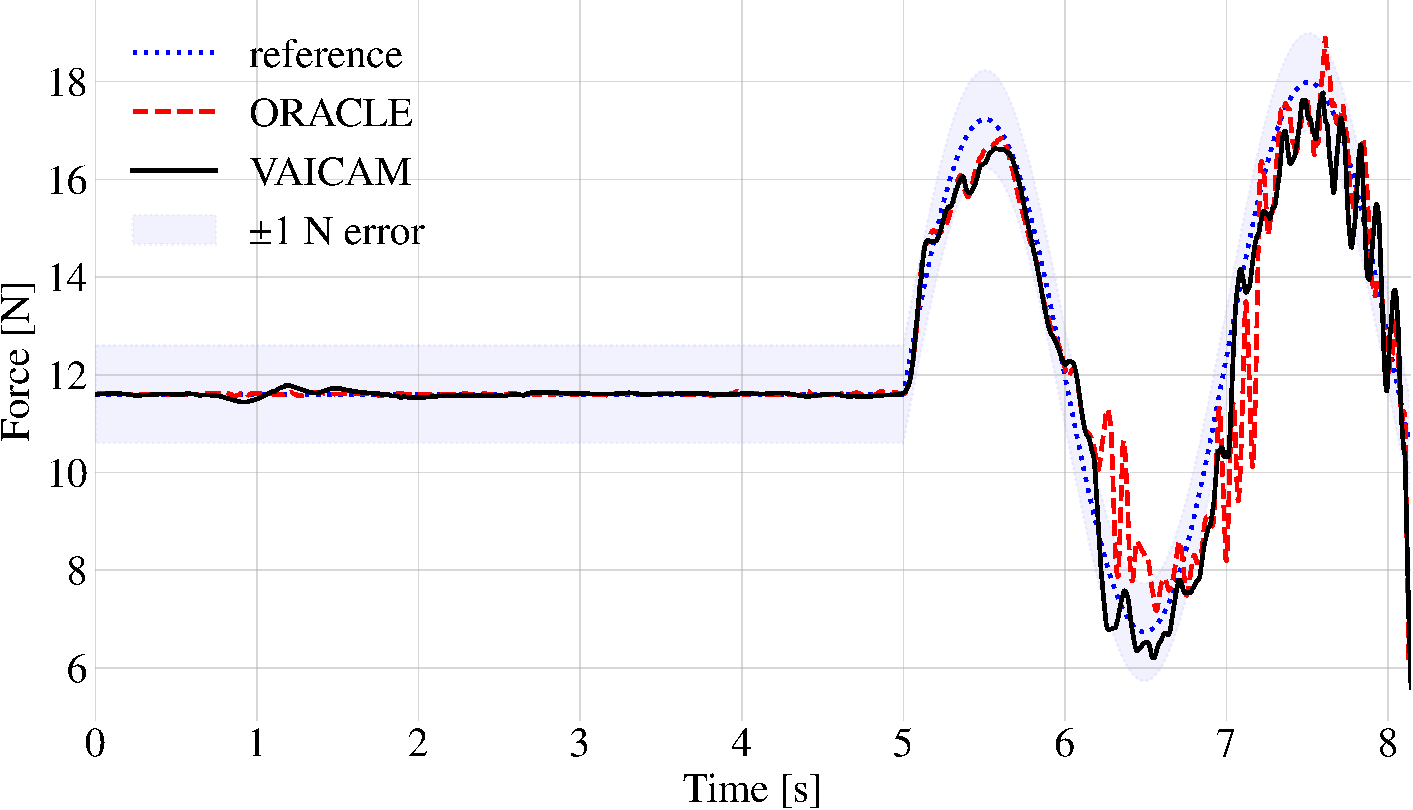}
\caption{Force tracking comparison between ORACLE \cite{petrone_optimized_2025} and VAICAM (ours) at $v = \SI{0.2}{\meter\per\second}$}.
\label{fig:oracle_vs_vaicam}
\end{figure}

As can be seen from Fig.~\ref{fig:oracle_vaicam_dfc_mse_sim}, the average force-tracking RMSE for VAICAM is approximately constant for $v \in \left[0, 0.35\right]\,\si{\meter\per\second}$, and starts increasing for $v > \SI{0.35}{\meter\per\second}$.
Thanks to the exploitation of the novel DMA, VAICAM outperforms both ORACLE and the base DFC.
A sample trajectory is considered in Fig.~\ref{fig:oracle_vs_vaicam}, visually showing that ORACLE's uncertainty at moderate velocities (due to the limited SMA) translates into worse force-tracking capabilities.
Furthermore, Fig.~\ref{fig:oracle_vaicam_dfc_mse_sim} shows that VAICAM also enhances the DFC performance, although with a lower improvement factor compared to ORACLE.
Lastly, it is worth noticing that VAICAM's RMSE standard deviation is lower than DFC's, thus manifesting a superior robustness \wrt the experimental conditions.

To confirm the improvement of VAICAM over both DFC and ORACLE, Table \ref{table:results} displays the numerical results by reporting the average RMSE of these 3 strategies for each velocity.
As $\eta$ confirms, VAICAM surpasses the DFC by up to 31\% (at $v = \SI{0.35}{\meter\per\second}$), while the improvement over ORACLE at maximum experimental speed ($v = \SI{0.5}{\meter\per\second}$) reaches 127\%.

\begin{table}
\centering
\caption{Average RMSE of baseline DFC \cite{roveda_sensorless_2021}, ORACLE \cite{petrone_optimized_2025}, and VAICAM (ours) across trajectories --- $\eta$ indicates the improvement factor of VAICAM over DFC or ORACLE.}
\label{table:results}
\resizebox{1.0\columnwidth}{!}{
\begin{tabular}{|c|ccc|cc|} 
\hline
\textbf{Velocity} & \multicolumn{3}{c|}{\textbf{RMSE [\si{\newton}]}} & \multicolumn{2}{c|}{$\bm\eta$} \\
\textbf{[\si{\meter\per\second}]} & \textbf{DFC} & \textbf{ORACLE} & \textbf{VAICAM} & \textbf{DFC} & \textbf{ORACLE} \\ 
\hline
0.01 & 2.0579 & 2.1222 & 2.0676 & 0.9953 & 1.0264 \\
0.05 & 2.0602 & 2.1867 & 1.8917 & 1.0891 & 1.1559 \\
0.1 & 2.1022 & 2.2118 & 2.0259 & 1.0377 & 1.0918 \\
0.15 & 2.1563 & 3.135 & 2.0528 & 1.0504 & 1.5272 \\
0.2 & 2.199 & 3.5081 & 2.0697 & 1.0624 & 1.695 \\
0.25 & 2.3638 & 3.8424 & 2.0362 & 1.1609 & 1.887 \\
0.3 & 2.4995 & 4.3858 & 2.1098 & 1.1847 & 2.0788 \\
0.35 & 2.6919 & 4.6601 & 2.0553 & 1.3097 & 2.2673 \\
0.4 & 2.973 & 4.883 & 2.3837 & 1.2472 & 2.0485 \\
0.45 & 3.1873 & 4.978 & 2.5117 & 1.269 & 1.982 \\
0.5 & 3.3923 & 5.4201 & 2.8123 & 1.2062 & 1.9273 \\
\hline
\end{tabular}
}
\end{table}

\section{\uppercase{Conclusions}}
\label{sec:conclusion}

This paper introduced VAICAM, a controller improving the state-of-the-art of robot-environment interaction tasks that require dynamic trajectories.
It makes use of an FFNN ensemble that acts as a model approximator considering the tangential velocity of the EE.
The controller is based on an impedance controller to ensure a compliant behavior, and a DFC to guarantee the desired force tracking characteristics.
An ensemble of FFNNs is developed, trained, and tested along the work.
The networks are able to predict the force that the manipulator will exert on the environment, allowing the computation of an optimal residual action. The latter is then added to the action of a baseline DFC to improve its force-tracking capabilities, by correcting the next commanded position.
Tested in a simulated environment, VAICAM outperformed the baseline DFC \cite{roveda_sensorless_2021} and a similar algorithm from recent literature \cite{petrone_optimized_2025}.

Possible future improvements of the proposed strategy are:
\begin{enumerate*}[label=(\roman*)]
    \item extending the comparison with other controllers from the literature, \eg \cite{iskandar_hybrid_2023};
    \item assessing whether these controllers can benefit from a MA like the one developed in this work to improve their performance;
    \item deploying the algorithm on real hardware, for a complete validation of VAICAM's performance;
    \item extending the state space to include tangential position components, to tackle variable-stiffness environments;
    \item including the DMA in optimal planning algorithms foreseeing robot-environment interaction trajectories \cite{petrone_time-optimal_2022}. 
\end{enumerate*}

\bibliographystyle{apalike}
{\small
\bibliography{IEEEabrv,OtherAbbrv,icinco-2025-vaicam}

\begin{thebibliography}{}

\bibitem[Agarap, 2019]{agarap_deep_2018}
Agarap, A.~F. (2019).
\newblock Deep {Learning} using {Rectified} {Linear} {Units} ({ReLU}).
\newblock {\em arXiv preprint: 1803.08375}.

\bibitem[Caccavale et~al., 1999]{caccavale_six-dof_1999}
Caccavale, F., Natale, C., Siciliano, B., and Villani, L. (1999).
\newblock Six-{DOF} impedance control based on angle/axis representations.
\newblock {\em {IEEE} Trans. Robot. Automat.}, 15(2):289--300.

\bibitem[Chua et~al., 2018]{chua_deep_2018}
Chua, K., Calandra, R., McAllister, R., and Levine, S. (2018).
\newblock Deep {Reinforcement} {Learning} in a {Handful} of {Trials} using {Probabilistic} {Dynamics} {Models}.
\newblock In {\em Adv. Neural Inform. Process. Syst.}, volume~32, pages 4754--4765.

\bibitem[Duan et~al., 2018]{duan_adaptive_2018}
Duan, J., Gan, Y., Chen, M., and Dai, X. (2018).
\newblock Adaptive variable impedance control for dynamic contact force tracking in uncertain environment.
\newblock {\em Robot. Auton. Syst.}, 102:54--65.

\bibitem[Featherstone and Orin, 2016]{featherstone_dynamics_2016}
Featherstone, R. and Orin, D.~E. (2016).
\newblock Dynamics.
\newblock In Siciliano, B. and Khatib, O., editors, {\em Springer {Handbook} of {Robotics}}, volume~3, pages 195--211. Springer, 2 edition.

\bibitem[Formenti et~al., 2022]{formenti_improved_2022}
Formenti, A., Bucca, G., Shahid, A.~A., Piga, D., and Roveda, L. (2022).
\newblock Improved impedance/admittance switching controller for the interaction with a variable stiffness environment.
\newblock {\em Compl. Eng. Syst.}, 2(3).
\newblock {A}rt. no. 12.

\bibitem[Haddadin et~al., 2022]{haddadin_franka_2022}
Haddadin, S., Parusel, S., Johannsmeier, L., Golz, S., Gabl, S., Walch, F., Sabaghian, M., Jähne, C., Hausperger, L., and Haddadin, S. (2022).
\newblock The {Franka} {Emika} {Robot}: {A} {Reference} {Platform} for {Robotics} {Research} and {Education}.
\newblock {\em {IEEE} Robot. Automat. Mag.}, 29(2):46--64.

\bibitem[Huang et~al., 2022]{huang_robust_2022}
Huang, H., Guo, Y., Yang, G., Chu, J., Chen, X., Li, Z., and Yang, C. (2022).
\newblock Robust {Passivity}-{Based} {Dynamical} {Systems} for {Compliant} {Motion} {Adaptation}.
\newblock {\em {IEEE/ASME} Trans. Mechatron.}, 27(6):4819--4828.

\bibitem[Iskandar et~al., 2023]{iskandar_hybrid_2023}
Iskandar, M., Ott, C., Albu-Schäffer, A., Siciliano, B., and Dietrich, A. (2023).
\newblock Hybrid {Force}-{Impedance} {Control} for {Fast} {End}-{Effector} {Motions}.
\newblock {\em IEEE Robot. Automat. Lett.}, 8(7):3931--3938.

\bibitem[Jung et~al., 2004]{jung_force_2004}
Jung, S., Hsia, T. C.~S., and Bonitz, R.~G. (2004).
\newblock Force {Tracking} {Impedance} {Control} of {Robot} {Manipulators} {Under} {Unknown} {Environment}.
\newblock {\em {IEEE} Trans. Contr. Syst. Technol.}, 12(3):474--483.

\bibitem[Khatib, 1987]{khatib_unified_1987}
Khatib, O. (1987).
\newblock A unified approach for motion and force control of robot manipulators: {The} operational space formulation.
\newblock {\em {IEEE} J. Robot. Automat.}, 3(1):43--53.

\bibitem[Kingma and Ba, 2015]{kingma_adam_2015}
Kingma, D.~P. and Ba, J. (2015).
\newblock Adam: {A} {Method} for {Stochastic} {Optimization}.
\newblock In {\em Int. Conf. Learn. Represent.}

\bibitem[Koenig and Howard, 2004]{koenig_design_2004}
Koenig, N. and Howard, A. (2004).
\newblock Design and {Use} {Paradigms} for {Gazebo}, an {Open}-{Source} {Multi}-{Robot} {Simulator}.
\newblock In {\em IEEE Int. Conf. Intell. Robots Syst.}, volume~3, pages 2149--2154.

\bibitem[Li et~al., 2023]{li_adaptive_2023}
Li, K., He, Y., Li, K., and Liu, C. (2023).
\newblock Adaptive fractional-order admittance control for force tracking in highly dynamic unknown environments.
\newblock {\em Int. J. Robot. Res. Applic.}, 50(3):530--541.

\bibitem[Matschek et~al., 2023]{matschek_safe_2023}
Matschek, J., Bethge, J., and Findeisen, R. (2023).
\newblock Safe {Machine}-{Learning}-{Supported} {Model} {Predictive} {Force} and {Motion} {Control} in {Robotics}.
\newblock {\em {IEEE} Trans. Contr. Syst. Technol.}, 31(6):2380--2392.

\bibitem[Nagabandi et~al., 2018]{nagabandi_neural_2018}
Nagabandi, A., Kahn, G., Fearing, R.~S., and Levine, S. (2018).
\newblock Neural {Network} {Dynamics} for {Model}-{Based} {Deep} {Reinforcement} {Learning} with {Model}-{Free} {Fine}-{Tuning}.
\newblock In {\em IEEE Int. Conf. Robot. Automat.}, pages 7559--7566.

\bibitem[Paszke et~al., 2019]{paszke_pytorch_2019}
Paszke, A., Gross, S., Massa, F., Lerer, A., Bradbury, J., Chanan, G., Killeen, T., Lin, Z., Gimelshein, N., Antiga, L., Desmaison, A., Köpf, A., Yang, E., DeVito, Z., Raison, M., Tejani, A., Chilamkurthy, S., Steiner, B., Fang, L., Bai, J., and Chintala, S. (2019).
\newblock {PyTorch}: {An} {Imperative} {Style}, {High}-{Performance} {Deep} {Learning} {Library}.
\newblock In {\em Adv. Neural Inform. Process. Syst.}, volume~33, pages 8026--8037.

\bibitem[Petrone et~al., 2022]{petrone_time-optimal_2022}
Petrone, V., Ferrentino, E., and Chiacchio, P. (2022).
\newblock Time-{Optimal} {Trajectory} {Planning} {With} {Interaction} {With} the {Environment}.
\newblock {\em IEEE Robot. Automat. Lett.}, 7(4):10399--10405.

\bibitem[Petrone et~al., 2025]{petrone_optimized_2025}
Petrone, V., Puricelli, L., Pozzi, A., Ferrentino, E., Chiacchio, P., Braghin, F., and Roveda, L. (2025).
\newblock Optimized {Residual} {Action} for {Interaction} {Control} with {Learned} {Environments}.
\newblock {\em {IEEE} Trans. Contr. Syst. Technol.}
\newblock {Accepted} for publication.

\bibitem[Quigley et~al., 2009]{quigley_ros_2009}
Quigley, M., Conley, K., Gerkey, B., Faust, J., Foote, T., Leibs, J., Wheeler, R., and Ng, A. (2009).
\newblock {ROS}: an open-source {Robot} {Operating} {System}.
\newblock In {\em IEEE Int. Conf. Robot. Automat.}, volume~3.

\bibitem[Roveda et~al., 2020]{roveda_control_2020}
Roveda, L., Castaman, N., Franceschi, P., Ghidoni, S., and Pedrocchi, N. (2020).
\newblock A {Control} {Framework} {Definition} to {Overcome} {Position}/{Interaction} {Dynamics} {Uncertainties} in {Force}-{Controlled} {Tasks}.
\newblock In {\em IEEE Int. Conf. Robot. Automat.}, pages 6819--6825.

\bibitem[Roveda and Piga, 2021]{roveda_sensorless_2021}
Roveda, L. and Piga, D. (2021).
\newblock Sensorless environment stiffness and interaction force estimation for impedance control tuning in robotized interaction tasks.
\newblock {\em Auton. Robots}, 45(3):371--388.

\bibitem[Shen et~al., 2022]{shen_fuzzy-based_2022}
Shen, Y., Lu, Y., and Zhuang, C. (2022).
\newblock A fuzzy-based impedance control for force tracking in unknown environment.
\newblock {\em J. Mech. Sci. Technol.}, 36(10):5231--5242.

\bibitem[Shu et~al., 2021]{shu_adaptive_2021}
Shu, X., Ni, F., Min, K., Liu, Y., and Liu, H. (2021).
\newblock An {Adaptive} {Force} {Control} {Architecture} with {Fast}-{Response} and {Robustness} in {Uncertain} {Environment}.
\newblock In {\em Int. Conf. Robot. Biom.}, pages 1040--1045.

\bibitem[Siciliano and Villani, 1999]{siciliano_indirect_1999}
Siciliano, B. and Villani, L. (1999).
\newblock Indirect {Force} {Control}.
\newblock In {\em Robot {Force} {Control}}, pages 31--64. Springer US.

\bibitem[Yu et~al., 2024]{yu_adaptive_2024}
Yu, X., Liu, S., Zhang, S., He, W., and Huang, H. (2024).
\newblock Adaptive {Neural} {Network} {Force} {Tracking} {Control} of {Flexible} {Joint} {Robot} {With} an {Uncertain} {Environment}.
\newblock {\em {IEEE} Trans. Ind. Electron.}, 71(6):5941--5949.

\end{thebibliography}
}

\end{document}